%% file: root.tex
\documentclass[conference]{IEEEtran}
\usepackage{times}

\usepackage[numbers,sort]{natbib}
\usepackage{multicol}
\usepackage[bookmarks=true]{hyperref}

\usepackage{fancyhdr}  
\usepackage{amsmath} 
\usepackage{amssymb}  

\usepackage{booktabs}
\usepackage{hyperref}
\usepackage{dblfloatfix}
\usepackage{graphicx} 
\usepackage{caption}
\usepackage{subcaption}

\newcommand{\proc}[1]{\textsc{#1}}
\newcommand{\kw}[1]{\textbf{#1}}

\newcommand{\True}[0]{\kw{True}}

\newcommand{\None}[0]{\kw{None}}

\newcommand{\pddl}[1]{{\texttt{#1}}} 

\newcommand{\param}[1]{{\textit{?#1}}} 

\usepackage{xcolor}

\definecolor{MyDarkBlue}{rgb}{0,0.08,0.8}
\definecolor{MyDarkGreen}{RGB}{45,155,45}
\definecolor{MyDarkRed}{rgb}{0.8,0.02,0.02}
\definecolor{MyDarkOrange}{rgb}{0.40,0.2,0.02}
\definecolor{MyPurple}{RGB}{111,0,255}
\definecolor{MyRed}{rgb}{0.8,0.0,0.0}
\definecolor{MyGold}{rgb}{0.75,0.6,0.12}
\definecolor{MyDarkgray}{rgb}{0.66, 0.66, 0.66}
\definecolor{JiayuanColor}{rgb}{0.60,0.43,0.48}

\makeatletter
\newcommand\predicate[2]{\pddl{#1}(#2)} 
\newcommand\action[2]{\pddl{#1}(#2)} 
\makeatother

\usepackage{xcolor}
\usepackage{algorithm}
\usepackage[noend]{algpseudocode}
\usepackage[export]{adjustbox}
\algnewcommand\algorithmicdeclare{\textbf{Assume:}}
\algnewcommand\Declare{\item[\algorithmicdeclare]}

\usepackage{listings}
\let\labelindent\relax
\usepackage{enumitem}

\begin{document}

\title{Sequence-Based Plan Feasibility Prediction \\
for Efficient Task and Motion Planning}

\newcommand\blfootnote[1]{%
  \begingroup
  \renewcommand\thefootnote{}\footnote{#1}%
  \addtocounter{footnote}{-1}%
  \endgroup
}

\author{\authorblockN{Zhutian Yang$^{1,*}$,
Caelan Reed Garrett$^{2}$,
Tomás Lozano-Pérez$^{1}$, 
Leslie Pack Kaelbling$^{1}$,
Dieter Fox$^{2}$}
\authorblockA{$^{1}$Massachusetts Institute of Technology, $^{2}$NVIDIA Research}}


\setcounter{figure}{1}
\makeatletter
\let\@oldmaketitle\@maketitle
\renewcommand{\@maketitle}{\@oldmaketitle
  \begin{center}
    \includegraphics[width=\linewidth]{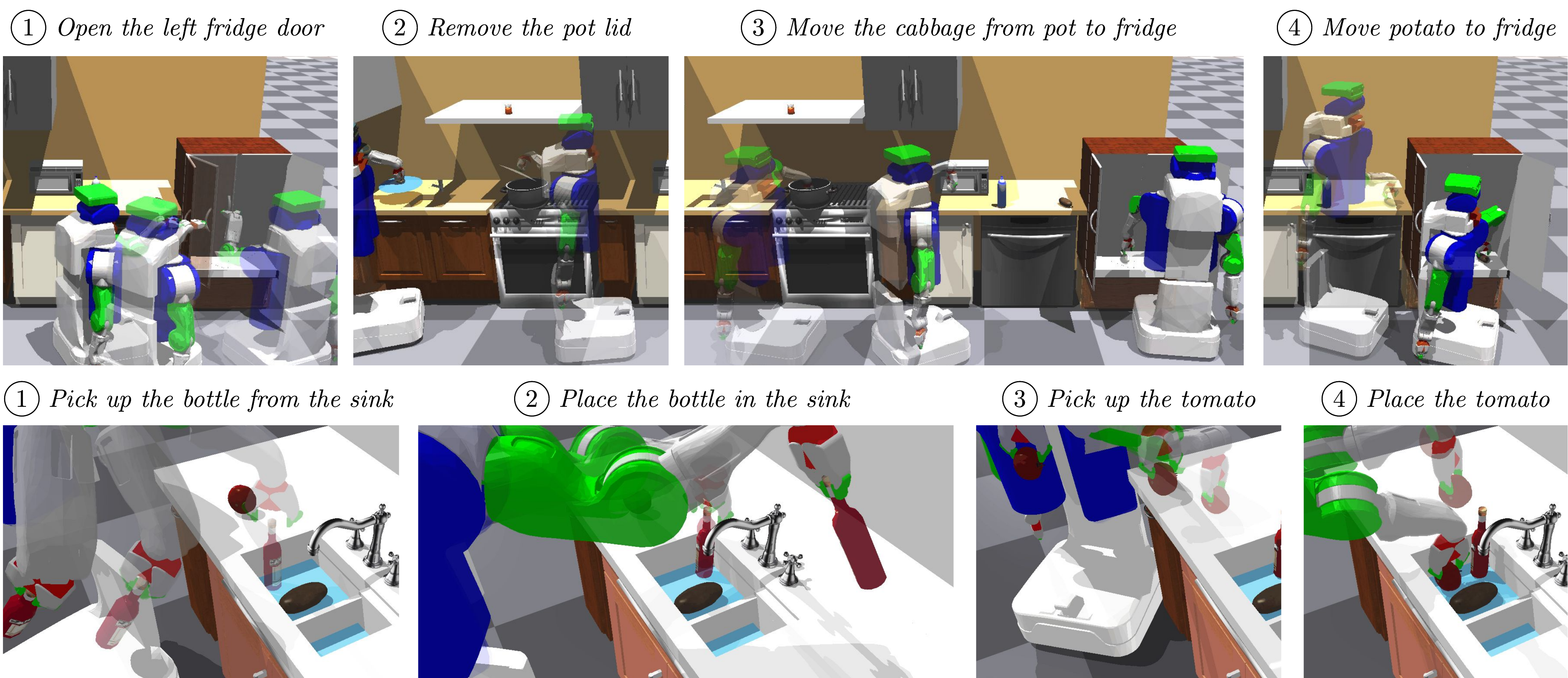} 
    \label{fig:overview}
  \end{center}
  \vspace{-12pt}
  \footnotesize{\textbf{Fig.~\thefigure:\label{fig:overview}}~
  Example solutions to two task and motion planning problems with complex geometric constraints, for which our PIGINet reduces the planning time by filtering out infeasible task plans considered during planning. 
  \textbf{Top:} The goal is for all food to be in the fridge. \textbf{Bottom:} The goal is for the tomato to be in the sink. Directly picking and placing the goal objects in both problems is not possible due to obstruction caused by articulated and movable obstacles. }
  \medskip}
\makeatother

\maketitle


\begin{abstract}
We present a learning-enabled Task and Motion Planning (TAMP) algorithm for solving mobile manipulation problems in environments with many articulated and movable obstacles. Our idea is to bias the search procedure of a traditional TAMP planner with a learned plan feasibility predictor. The core of our algorithm is PIGINet, a novel Transformer-based learning method that takes in a task plan, the goal, and the initial state, and predicts the probability of finding motion trajectories associated with the task plan. We integrate PIGINet within a TAMP planner that generates a diverse set of high-level task plans, sorts them by their predicted likelihood of feasibility, and refines them in that order. 
We evaluate the runtime of our TAMP algorithm on seven families of kitchen rearrangement problems, comparing its performance to that of non-learning baselines. 
Our experiments show that PIGINet substantially improves planning efficiency, cutting down runtime by 80\% on problems with small state spaces and 10\%-50\% on larger ones, after being trained on only 150-600 problems. Finally, it also achieves zero-shot generalization to problems with unseen object categories thanks to its visual encoding of objects. Project page \url{https://piginet.github.io/}.
\end{abstract}
\blfootnote{$^*$Work done during an internship at NVIDIA Research}
\IEEEpeerreviewmaketitle

\section{Introduction}
Planning for long-horizon robotic behavior in complex environments requires quick reasoning about the impact of the environment's geometry on what high-level plans are feasible. Many task and motion planning (TAMP)~\cite{Garrett2021, ortiz20222conflict} algorithms accomplish this by balancing the computational time spent on two processes. One is \textit{task planning}: finding high-level task plans consisting of discrete arguments that achieve the logical conditions specified by the goal. The other is \textit{motion planning}: generating 
continuous motion trajectories that are collision-free 
using sampling or optimization. 

Balancing between task planning and motion planning is particularly challenging when manipulable obstacles impose additional geometric constraints that makes it hard to find collision-free object placements or arm trajectories. For example, a mobile robot may be tasked with rearranging food items among fridges, cooking pots, and sinks. Doors and other food items may be blocking all paths that reach the goal objects or placement regions, as in the problems shown in Figure 1.
In these problems, the number of infeasible candidate task plans increases exponentially as the planning horizon and number of objects both increase.
An uninformed TAMP algorithm would waste a substantial amount of time attempting to satisfy many unsatisfiable constraints associated with infeasible task plans, {\it e.g.} by 
attempting to solve unsolvable motion planning subproblems, before working on the feasible ones. 

Some manipulation approaches deal with obstruction caused by storage units and containers by assuming that there exist predicates like (\texttt{reachable} \textit{sink}) or (\texttt{opened} \textit{door}) to help eliminate infeasible task plans during the discrete search process. However, this type of discretization of geometric state is not applicable to real-world situations where regions can be partially occupied, doors can be half open, and multiple doors need to be opened in order to enable reachability. In this paper, we propose an alternative strategy that avoids discretization by generating a diverse set of task plans and pruning out the infeasible task plans using a neural network.



At the core of our framework is \textit{\textbf{PIGINet}}, a plan feasibility prediction network based on the transformer architecture. Given a candidate task \textit{\textbf{P}}lan with \textit{\textbf{I}}mage features of objects, 2) the \textit{\textbf{G}}oal formula, and 3) the relations and continuous values in the \textit{\textbf{I}}nitial state, PIGINet outputs a probability that the task plan is feasible. The elements of each action or relation in the initial state--such as text, object poses, and door joint angles--are processed to produce embeddings of the same dimension and fused together to produce each token in the input sequence to transformer encoder. A pre-trained CLIP model \cite{radford2021learning} is used to generate corresponding text and image embeddings. For each distribution of tasks, the model is trained with up to 4000 plans and plans for up to 600 environments. 

We deploy our trained model in a TAMP algorithm that generates a large number of task plans, sorts them by the model's predicted likelihoods of feasibility, and refines them in order of feasibility until a solution is found.
We evaluate the success rate and runtime of our learning-enabled TAMP algorithm on unseen problem instances in comparison to a baseline and ablations. Our experiments show that learning to predict task plan feasibility can substantially improve planning performance, cutting down runtime by 80\% on problems with a low dimensional state space and 10\%-50\% on larger ones. It also achieves zero-shot generalization to problems with unseen object categories thanks to its visual encoding of objects.


\section{RELATED WORK}

Our method builds on prior work in task and motion planning (TAMP), learning to expedite TAMP, and sequence prediction for robot manipulation.

\paragraph{Task and Motion Planning} 
One approach to TAMP first performs a search over high-level task plans and then refines each plan to be refined using sampling or optimization~\cite{Garrett2021}. 
Two of these algorithms~\cite{ren2021extended,ortiz2022diverse} use {\em diverse planning} techniques~\cite{katz2020top}, which identify multiple distinct plans to produce candidate task plans.
Because diverse planners are unaware of the geometry of the world, many candidate plans have the same sources of infeasibility, \textit{e.g.,} due to unreachable objects or regions. 
As a result, these TAMP algorithms waste time finding continuous values for similar, unsatisfiable motion planning problems associated with those task plans.
Many TAMP algorithms contain mechanisms that provide specific feedback~\cite{Garrett2021} to the search over task plan using 
failed motion queries~\cite{dantam2018incremental} 
or unsatisfiable constraint sets~\cite{garrett2021sampling}.
However, these approaches require expensive geometric planning to first identify failed task plans and second generate feedback.


\paragraph{Learning to speed up TAMP} TAMP slows down exponentially as the problem horizon and the number of manipulable objects increase. Silver {\it et al.}~\cite{silver2021planning} developed a Graph Neural Network (GNN) approach that ignores irrelevant objects in table-top rearrangement problems, taking into account object obstruction. 
Khodeir {\it et al.}~\cite{khodeir2021learning} extended this approach to predict which
samplers should be prioritized when solving for continuous values. Kim {\it et al.}~\cite{kim2020learning} learned a cost-to-go heuristic estimator using a relational embedding of the state to guide search. 
Several learning-for-TAMP approaches learn single-action feasibility classifiers using object poses and relative distances~\cite{wells2019learning}, depth image~\cite{driess2020deep, xu2022accelerating, bouhsain2022learning}, or point cloud~\cite{park2022scalable} encoding of the environment.
Compared to work on action feasibility, our approach of classifying feasibility of entire plans enables us to 1) discard infeasible task plans without ever performing motion planning and 2) consider constraints arising from actions late in the plan that restrict choices early in the plan (e.g. whether two doors need to be opened depends on how many objects must be placed inside later and how large they are). Also, we are leveraging, instead of replacing, task planning \cite{driess2021learning, khodeir2021learning}, by providing the learner with sound task plans, easing the learning burden and, in practice, greatly expanding the learner’s ability to generalize to varied initial states, goals, and even actions. Furthermore, we are deploying powerful pre-trained language and vision models to represent complex scenes in order to learn models that generalize well from relatively small amounts of training data (150-600 problems for each problem set).



\paragraph{Sequence-based modeling for robotic manipulation} We are inspired by recent works that used attention-based neural networks to encode the state, fusing multi-modal inputs, and making object-centric decisions. 
Zhu {\it et al.}~\cite{zhu2021hierarchical} generate the next action name and action parameters by encoding the states as symbolic and geometric scene graphs and then process them with GNNs. 
Liu {\it et al.} ~\cite{liu2022structformer} predict object poses for semantically meaningful arrangements by encoding text and point-cloud embeddings in the same sequence for a transformer encoder.
Blukis {\it et al.} ~\cite{blukis2022persistent} generate subgoals given language instructions by encoding a sequence of previous subgoals and combining them with a language embedding. 
Yuan {\it et al.}~\cite{yuan2022sornet} learn object embeddings by encoding a sequence consisting of image patches of the whole scene and canonical views of each object in a transformer architecture and then use the learned embedding for querying object spatial relations or a direction for gripper movement. Our work uses similar techniques for merging multi-modal input but deals with more complex spatial relationships between objects, reasoning about obstruction while also dealing with extraneous inputs due to the large state space of our mobile manipulation problems.

\section{PROBLEM FORMULATION}

We represent TAMP problems using an extension of the Planning Domain Definition Language (PDDL), a logic-based action language, that supports planning with continuous values~\cite{garrett2020PDDLStream}. We define a TAMP {\em domain} $\langle \mathcal{P}, \mathcal{C}, \mathcal{A}\rangle$ by a set of {\em predicates} $\mathcal{P}$, {\em constants} $\mathcal{C}$ and {\em actions} $\mathcal{A}$. {\em Predicates} and {\em actions} can be represented as tuples consisting of a name and a list of typed arguments. The arguments may be (1) discrete, such as object and part names, or (2) continuous, such as object poses, object grasps, robot configurations, object joint angles, and robot trajectories. {\em Constants} name objects that are useful by all problems in the domain, such as object categories.


A TAMP {\em problem} $\langle \mathcal{O}, \mathcal{I}, \mathcal{G}, \mathcal{A}\rangle$ is defined by a set of {\em objects} $\mathcal{O}$, a set of {\em initial literals} $\mathcal{I}$, and a conjunctive set of {\em goal literals} $\mathcal{G}$. A {\em literal} is a predicate with an assignment of values to its arguments. The set of initial literals defines a state of the world. Each {\em grounded action} defines a deterministic transition of the world state.
Table~\ref{tab:example} shows some example literals of each construct. Note that ``fridge1:door1'', ``fridge1:space1'', and ``fridge1'' are three different objects in the planning problem since they afford different actions. 

{
\begin{table}[tp]
\centering\footnotesize
\begin{tabular}{llll} 
\toprule
$\mathcal{I}nit$                                & $\mathcal{G}oal$            \\ 
\toprule
\predicate{IsJointTo}{door3, fridge1}      & \predicate{Closed}{door3}          \\
\texttt{IsType}(tomato1, \texttt{@food}) & \predicate{In}{tomato1, storagespace2}     \\
\predicate{Supported}{tomato1, table2, $p_0$}             & \predicate{On}{tomato1, table2}     \\
\bottomrule
\end{tabular}
\caption{Example initial facts ($\mathcal{I}$) and goal conditions ($\mathcal{G}$), where \texttt{@} indicates constants.} 
\label{tab:example}
\end{table}
}


{
\begin{table}[tp]
\centering\footnotesize
\begin{tabular}{llll} 
\toprule
$\textit{Plan skeleton}$ $\hat{\pi}$         & $\textit{Task plan}$  $\pi$                 \\ 
\toprule
\action{pull}{fridge1:door1; $a_0$, \param{$a_1$}, \param{$g_1$}, \param{$t_1$}}         & $\langle$\texttt{pull}, fridge1:door1$\rangle$   \\
\action{pick}{tomato1; $p_0$, \param{$q_2$}, \param{$g_2$}, \param{$t_2$}}      & $\langle$\texttt{pick}, tomato1$\rangle$    \\
\bottomrule
\end{tabular}
\caption{An example plan skeleton and corresponding task $\mathcal{P}$lan used for plan feasibility prediction.  Symbols starting with \param{} are variables to be assigned during skeleton refinement. } 
\label{tab:plan}
\vspace{-15pt}
\end{table}

A {\em solution} is a finite sequence of grounded action instances that, when sequentially applied to the initial state $\mathcal{I}$, produces a terminal state where the goal literals $\mathcal{G}$ all hold. 
During the course of planning, many TAMP algorithms reason about {\em plan skeletons} $\hat{\pi}$, partial solutions where the continuous arguments are yet to be bound (denoted by the prefix \param{}).
Some skeletons can be turned into solutions through {\em refinement} by searching over continuous values for unbound parameters that satisfy the plan's constraints, such as inverse kinematics and collision-free constraints. 
However, successfully refining a skeleton is not always feasible.
A task plan $\pi$ is a skeleton without continuous arguments, as shown on the right of Table~\ref{tab:plan}. 

}


\begin{figure}[tp]
\begin{center}
\includegraphics[width=0.46\textwidth]{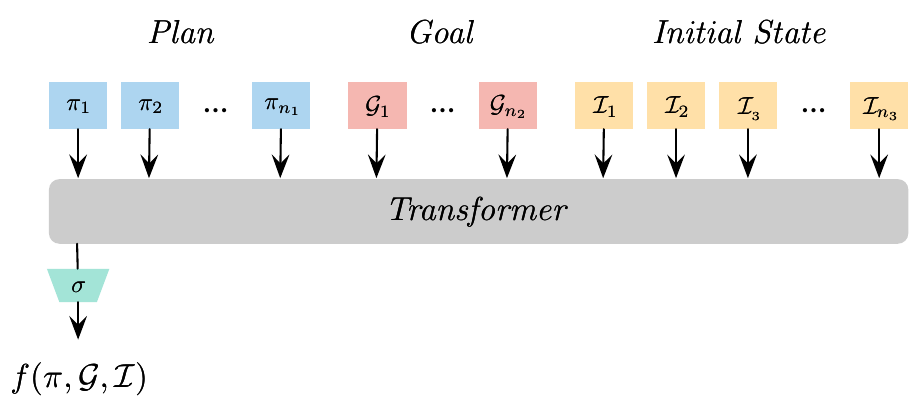}
\end{center}
\caption{The input and output of our plan feasibility predictor. \textbf{Input} is tokenized high-level actions, goal conditions, and facts in the initial state. \textbf{Output} include the predicted likelyhood that the high-level plan can be refined to produce motion trajecories to satisfy all geometric constraints.}
\label{fig:io}
\end{figure}

\begin{figure}[tp]
\begin{center}
\includegraphics[width=0.49\textwidth]{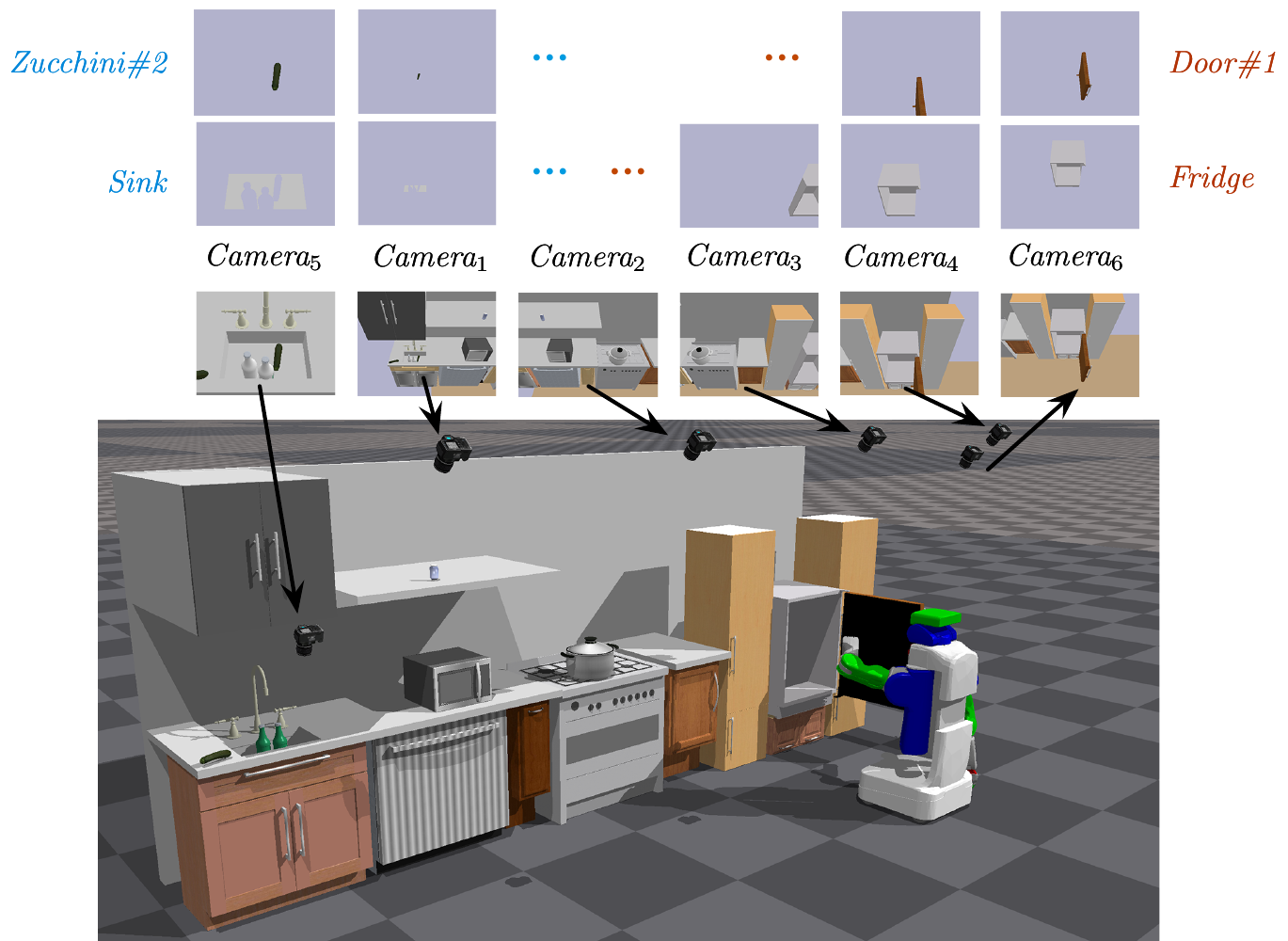}
\end{center}
\caption{Example camera poses and segmented images rendered in PyBullet. The images with only background color are omitted. \textit{Cameras} 1-4 form an array that covers the whole kitchen, while \textit{Cameras} 5-6 look at goal-related regions.}
\label{fig:cameras}
\end{figure}

Usually, task plans are produced and refined in order of plan length. Our idea is to refine them in order of plan feasibility, the likelihood that the refinement process can find a set of values for all continuous arguments in $\hat{\pi}$. The role of the plan feasibility predictor $f$ is to take in a task plan $\pi$, initial literals ${\cal I}$, and goal literals ${\cal G}$, and output a score $f({\cal I}, \pi, {\cal G})$, as shown in Figure~\ref{fig:io}.
We use $f$ generically as a scoring function for ranking a batch of task plans for refinement. 
Note that this work addresses traditional TAMP, where the system is assumed to be deterministic and observable.
Thus, the planner has full observability of the state, which includes information such as the pose of each object, even if it's occluded from the perspective to the camera or fully enclosed, as well as whether the object is supported by its initial container. 
Since the feasibility checker also has access to the full state, we render images from desired camera poses to take advantage of pre-trained computer vision networks that takes in RGB image input, as shown in Figure~\ref{fig:cameras}. Here, 6 image viewpoints are used because (a) the kitchen is wide and objects will be too small to be useful for feature extraction when the camera is far away and (b) we need cameras that are looking from top-down views (to see the empty area in a sink or pot) and side views (to see into cabinets and fridges).





\section{LEARNING METHOD}

\begin{figure}[tp]
\begin{center}
\includegraphics[width=0.5\textwidth]{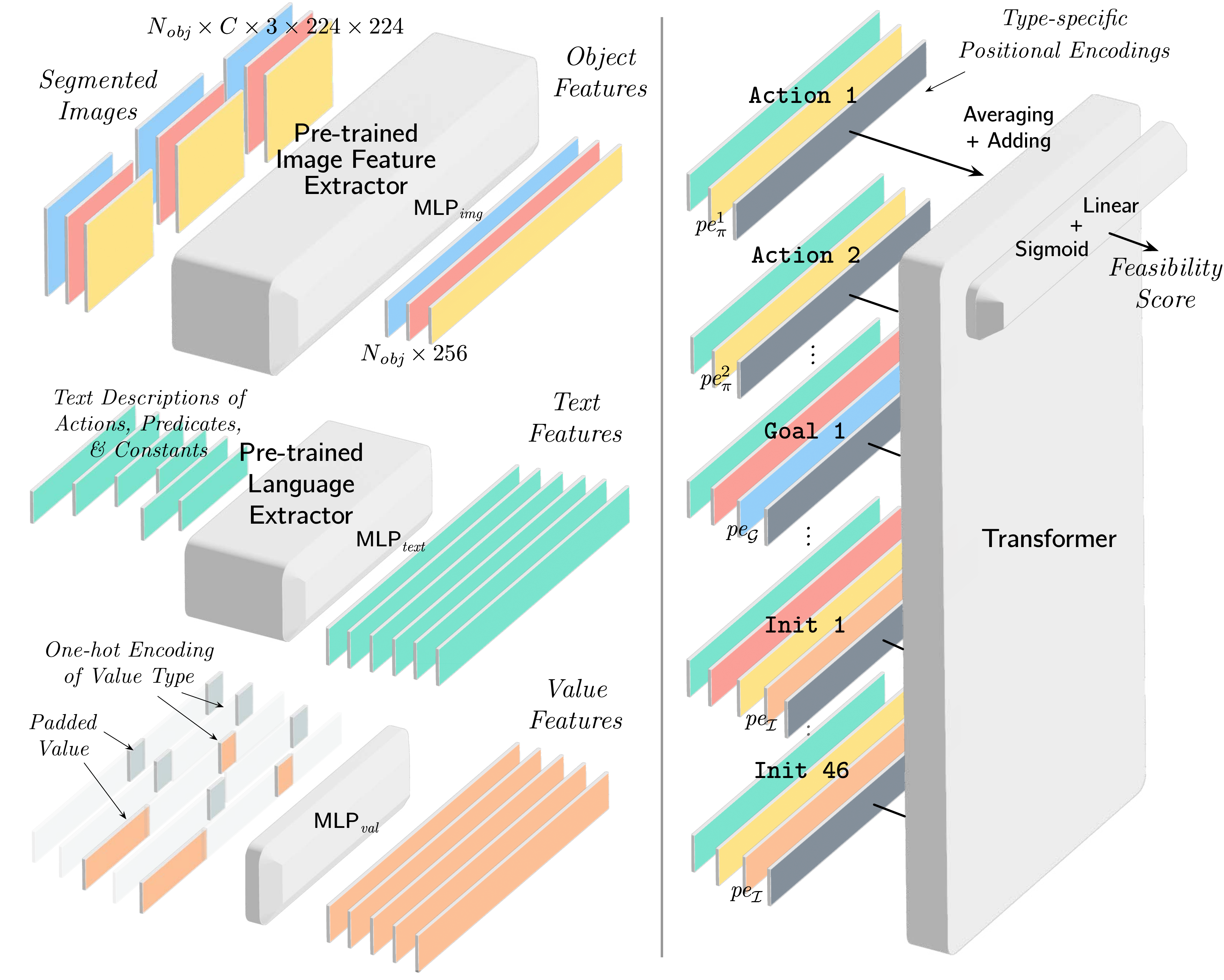}
\end{center}
\caption{PIGINet architecture. \textbf{Left}: Encoding of objects, values, and text into feature vectors of the same length. \textbf{Right}: Construction of input sequence for transformer encoder from lists of variable lengths from the plan, goal, and initial state.}
\label{fig:arch}
\vspace{-15pt}
\end{figure}

Given a TAMP problem and a task plan, PIGINet first builds a dictionary of embeddings for objects, continuous values, and text using type-specific encoders. Next, it converts each action in the plan, each literal in the goal, and the initial state into tokens and stack them to produce one input sequence for a transformer encoder. The decoded output is the probability of plan feasibility. The architecture is illustrated in Figure~\ref{fig:arch}. 

\subsection{Encoding objects, text, and continuous values}

Constructing the input sequences requires encoding strategies for elements of different modalities that form the initial state, goal, and candidate plan, including objects, texts, and continuous values.

\textbf{Objects} in the planning problem are represented by images. We use images as an approximation of the geometric state to leverage vision models. Although the planner has full state information including collision geometries, the task planning process normally would not consider geometry as it reasons over symbolic states. We 1) render RGB images of the scene from $C$ cameras and query instance segmentation of each object $\{\{x_{c,o}\}_{c=1}^C\}_{o=1}^{|O|}$, 2) extract their features with a pre-trained vision network $f_{\textit{img}}$, 3) concatenate the features from different cameras for each object, and 4) reduce the dimension of the resulting feature vectors with a three-layer Multi-Layer Perceptron (MLP), 
$$g_{\textit{img}}(o) = \textit{MLP}_{\textit{img}}([f_{\textit{img}}(x_{1, o}); ...; f_{\textit{img}}(x_{C, o})]).$$

\textbf{Text} used in the domain, which includes the names of predicates and actions as well as constants, is represented as feature vectors with the same dimension as object and value embeddings. There is a fixed number of such words in a domain, such as \texttt{SupportedBy}, \texttt{Pull}, and \texttt{@bottle}. For a word $w$, we first describe it using a colloquial English phrase or sentence $s$. Rephrasing helps the network deal with out-of-distribution names in the domain description file such as \texttt{isjointto}, which is rewritten to ``\textit{this is a joint of that object}''. Then, we encode it with a pre-trained language embedding network $f_{\textit{text}}$ and transform the output feature through a linear and a ReLU layer, 
$$g_{\textit{text}}(w) = \textit{MLP}_{\textit{text}}(f_{\textit{text}}(s)).$$

\textbf{Continuous values} in the initial literals, such as object poses $p = (x, y, z, \textit{yaw})$ and door joint angles $a = (\theta,)$, are each treated as typed tuples. We define a fixed set of value types $\mathcal{T}$ and their corresponding tuple length $\mathcal{L}$. First, we normalize the values to fall in the range $[-1, 1]$. For example, the $x, y, z$ values are normalized with respect to the bounding box of the world. The door angles, originally are in $[0, \mathcal{L}_{\textit{upper}}]$ where $0$ means the door is fully closed and $L_{\textit{upper}} \le \pi$, are normalized to $[0, 1]$. Then, we zero-pad tuples of varying lengths to become feature vectors of the same dimension $\sum_{i=0}^{|\mathcal{T}|} L_i$. For example, if value $v = (v_1, ..., v_{L_j})$ has type $\mathcal{T}_j$, the resulting vector $\Tilde{v}$ has zero in all positions except from $\sum_{i=0}^{j-1} \mathcal{L}_i$ to $\sum_{i=0}^{j} \mathcal{L}_i$. Then, we concatenate a one-hot encoding of $\mathcal{T}_j$ with $\Tilde{v}$. For example, if $p_0 = (x, y, z, \textit{yaw})$ and $a_0 = (\theta,)$ are the only values used in the initial literals, $\mathcal{T} = [1, 2]$, $\mathcal{L} = [4, 1]$, the processed features will be
\begin{align*}
p_0 &: [\underbrace{1, 0}_{\tau = 1, \text{ type is pose}}, \underbrace{\hat{x}, \hat{y}, \hat{z}, \hat{\textit{yaw}}}_{\textit{normalized value}}, 0], \\
\text{and } a_0 &: [\underbrace{0, 1}_{\tau = 2, \text{ type is angle}}, 0, 0, 0, 0, \underbrace{\hat{\theta}}_{\textit{normalized value}}].
\end{align*}

Lastly, we process the resulting feature vectors with a linear and a ReLU layer,
$$g_{\textit{val}}(v) = \textit{MLP}_{\textit{val}}([\textit{one-hot}(j, |\mathcal{T}|); \Tilde{v}]).$$


\subsection{Transformer encoder for Plan, Images, Goal, Init}

We view each literal in $\mathcal{I}nit$ and $\mathcal{G}oal$, as well as each action in $\pi$, as one list of multi-modal elements $\mathbf{z} = [z_1, ..., z_m]$ with element types $\mathbf{t} = [t_1, ..., t_m]$. For literals, $m \in [2, 4]$; for actions, $m \in [1, 2]$. We produce a token out of the list by mapping each element of the tuple to its feature embedding, averaging the embeddings, and then adding a \textit{positional encoding}. We choose to average the embeddings, as opposed to concatenating them, in order to handle tuples of varying length. In practice, the two aggregation methods yield similar training performance. \textit{Positional encodings} are used by transformer architectures as an addition to each input tokens because the order matters for sequence-based models. Each action in $\pi$ has a different sinusoidal positional encoding $pe_{\pi}^{1:k}$ where $k$ is the number of actions. For the set of literals in $\mathcal{G}$ and in $\mathcal{I}$, the sequence of constituting literals doesn't matter, so the positional encoding could denote their token type. All literals in $\mathcal{G}$ have the same learned positional encoding $pe_{\mathcal{G}}$, and all literals in $\mathcal{I}$ have the same learned positional encoding $pe_{\mathcal{I}}$. In summary, the process of translating a list $\mathbf{z}$ into an input token $x_k$ is as follows, where $g_{t_i} \in [g_\textit{img}, g_\textit{text}, g_\textit{val}] $ are defined previously and $\textit{type}(\mathbf{z}) \in [\pi, \mathcal{G}, \mathcal{I}]$,
$$x_k = h(\mathbf{z}, \mathbf{t}) = \frac{1}{m} \sum_{i=1}^m g_{t_i}(z_i) + pe_{\textit{type}(\mathbf{z})}^k.$$


The transformer encoder takes in the sequence of tokens $x$ representing $\pi$, $\mathcal{G}$, and $\mathcal{I}$,
$$x_{1:n} = [h(\pi_1), ..., h(\pi_{n_1}), h(\mathcal{G}_1), ..., h(\mathcal{G}_{n_2}), h(\mathcal{I}_1), ..., h(\mathcal{I}_{n_3})].$$

Embeddings of initial literals are dropped uniformly at random when the length of the sequence exceeds the max sequence length $n$. We use multi-headed attention to enable each position in the sequence to attend to others, except for within the plan tokens. The plan tokens use a causal mask, to build in the bias that the feasibility of one action doesn't depend on future actions. In practice, models trained with causal-plan mask and full mask achieve similar learning performance. Using three layers of residual attention blocks, we get an output $y$ with the same length as input $x$,
$$y_{1:n} = \textit{Transformer}(x_{1:n}).$$

We keep only the first position of the output and add linear and sigmoid layers to produce a nonnegative feasibility score,
$$f(\pi, \mathcal{G}, \mathcal{I}) = \textit{MLP}_{out}(y_1).$$

\subsection{Training}
We train PIGINet using the binary cross-entropy classification loss between the prediction and the label. We add a positive weight to the loss function according to the ratio of negative to positive training examples. The models are trained until prediction accuracy converges on the validation set. We use the confidence of the prediction as feasibility score. The whole architecture is trained end-to-end.



\section{PLANNING ALGORITHM}

We developed a new PDDLStream~\cite{garrett2020PDDLStream} algorithm, {\em batch-sorted}, that uses our PIGINet for feasibility prediction by feeding it a large number of candidate task plans for scoring. 
Algorithm~\ref{alg:batch} gives the pseudocode for the main algorithm \proc{batch-sorted-tamp}.
Similar to the {\em focused} algorithm~\cite{garrett2020PDDLStream}, it lazily instantiates a set of free action parameters $X$, that stand in for actual continuous values, using the available sampling operations. 
Namely, it recursively creates free parameters that optimistically represent the possible output of the sampling operations. 
The subroutine \proc{new-parameters} increases the depth of the recursive parameter instantiation.

\proc{batch-sorted-tamp} repeatedly searches for $k$ distinct plan skeletons to make up a single skeleton batch $\Pi_k$, where $k \geq 1$.
New plans are identified by performing a forward plan-space search $\proc{forbid-search}$ that forbids any previously identified plans $\Pi$ from the search's open list~\cite{garrett2020diverse}.
After $k$ attempts, the batched plans $\Pi_k$ are scored using the learned feasibility predictor $f(\pi, \mathcal{G}, \mathcal{I})$.
Plans that are predicted to have feasibility $<0.5$ are discarded. 
The rest are sorted in decreasing order of predicted likelihood of feasibility.
The algorithm attempts to refine each plan in order using sampling via \proc{refine-plan}.
This involves searching over combinations of sampler output values that bind each free parameter and jointly satisfy the preconditions of each step in the plan.
The first fully-bound plan $\pi_*$ is returned as a solution.

When running without a feasibility predictor ({\it i.e.} $f(\pi, \mathcal{G}, \mathcal{I}) = 1$), \proc{batch-sorted-tamp} is a {\em complete} algorithm for PDDLStream planning, assuming that \proc{forbid-search} is a complete discrete search.
This follows from the fact that the set of parameters $X$ defines a finite discrete search subproblem. 
\proc{forbid-search} will eventually enumerate all solutions to this problem.
Once exhausted, \proc{new-parameters} expands the subproblem, admitting more solutions.
We attempt to refine each solution using \proc{refine-plan}, which can be done indefinitely, for example, in a parallel thread.
When a feasibility predictor is present and also produces false negatives, we can obtain a complete algorithm by, rather than rejecting them, refining them at a lower computation rate.
Finally, because these algorithms are complete for PDDLStream, they are also {\em probabilistically complete} with respect to the underlying manipulation problem if the samplers satisfy some sample coverage properties~\cite{garrettIJRR2018}.


{\footnotesize
\begin{algorithm}[!h]
  \caption{Batch Sorted TAMP Plan Prediction} 
  \label{alg:batch}
  \begin{algorithmic}[1] 
    \Require Feasibility predictor: $f(\pi, \mathcal{G}, \mathcal{I}) \to [0, 1]$
    \Procedure{batch-sorted-tamp}{${\cal O}, \mathcal{I}, \mathcal{G}, \mathcal{A}; k$} 
        \State $X \gets \emptyset$ \Comment{Initialize plan free parameters}
        \State $\Pi \gets \emptyset$ \Comment{Initialize identified plans}
        \While{\True}
            \State $\Pi_k \gets \emptyset$ \Comment{Initialize batch of at most $k$ plans}
            \For{$i \in \{1, ..., k\}$} 
                \State $\pi \gets \proc{forbid-search}({\cal O}, X, {\cal I}, \mathcal{G}, \mathcal{A}; \Pi)$ 
                \If{$\pi \neq \None$} \Comment{Identified a new plan}
                    \State $\Pi \;\cup\!= \{\pi\}$
                    \State $\Pi_k  \;\cup\!= \{\pi\}$ 
                \Else \Comment{Infeasible: add more parameters}
                    \State $X \;\cup\!= \proc{new-parameters}({\cal O}, {\cal I}, X)$ 
                \EndIf
            \EndFor

            \State $P = \big\{\langle f(\pi, \mathcal{G}, \mathcal{I}), \pi\rangle \mid \pi \in \Pi_k\big\} $ 
            \For{$\langle p, \pi \rangle \in \kw{reversed}(\kw{sorted}(P))$} 
                \If{$p \geq 0.5$} \Comment{Filter plans}
                    \State $\pi_* = \proc{refine-plan}(\pi; \mathcal{I}, \mathcal{G})$ 
                    \If{$\pi_* \neq \None{}$} 
                        \State \Return $\pi_*$ \Comment{Return the bound solution}
                    \EndIf
                \EndIf
            \EndFor
       \EndWhile
       \EndProcedure
\end{algorithmic}
\end{algorithm}
}





\section{DATA GENERATION} 


We study a collection of tasks that require moving objects to target surfaces or storage units, using actions \texttt{move} (robot motion), \texttt{pick} and \texttt{place} (prehensile object manipulation), and \texttt{pull} (operating single degree-of-freedom mechanisms such as doors and drawers). We use \textit{problem set} to refer to a set of problems that are similar in object types, initial object relations, and goal formulas, but vary in scene layout, object assets, and initial world configuration. 

We first experiment with a simple setting, the \textit{Fridge (FG)} problems, where the scene contains a small set of objects so a single camera can capture all objects. Then we experiment in a larger setting, the \textit{Kitchen (KC)} problems, where the problems contain extraneous articulated and movable objects. Because the kitchens are wide, there are 6 simulated cameras: 4 arranged in a line pointing to the front face of the long kitchen space and 2 close-up cameras pointing at the sink and cooking pot from a top-down view. Together, they capture the shape and visibility of objects. In practice, we observe that models trained without images from the close-up cameras tend to perform not as well.

We briefly describe the problem sets as follows. For detailed descriptions of all seven problems and differences in PIGINet hyper-parameters for training on them, please see Appendix. 

\begin{figure*}[tp]
    \centering
    \begin{subfigure}[b]{0.49\textwidth}
        \includegraphics[width=\textwidth]{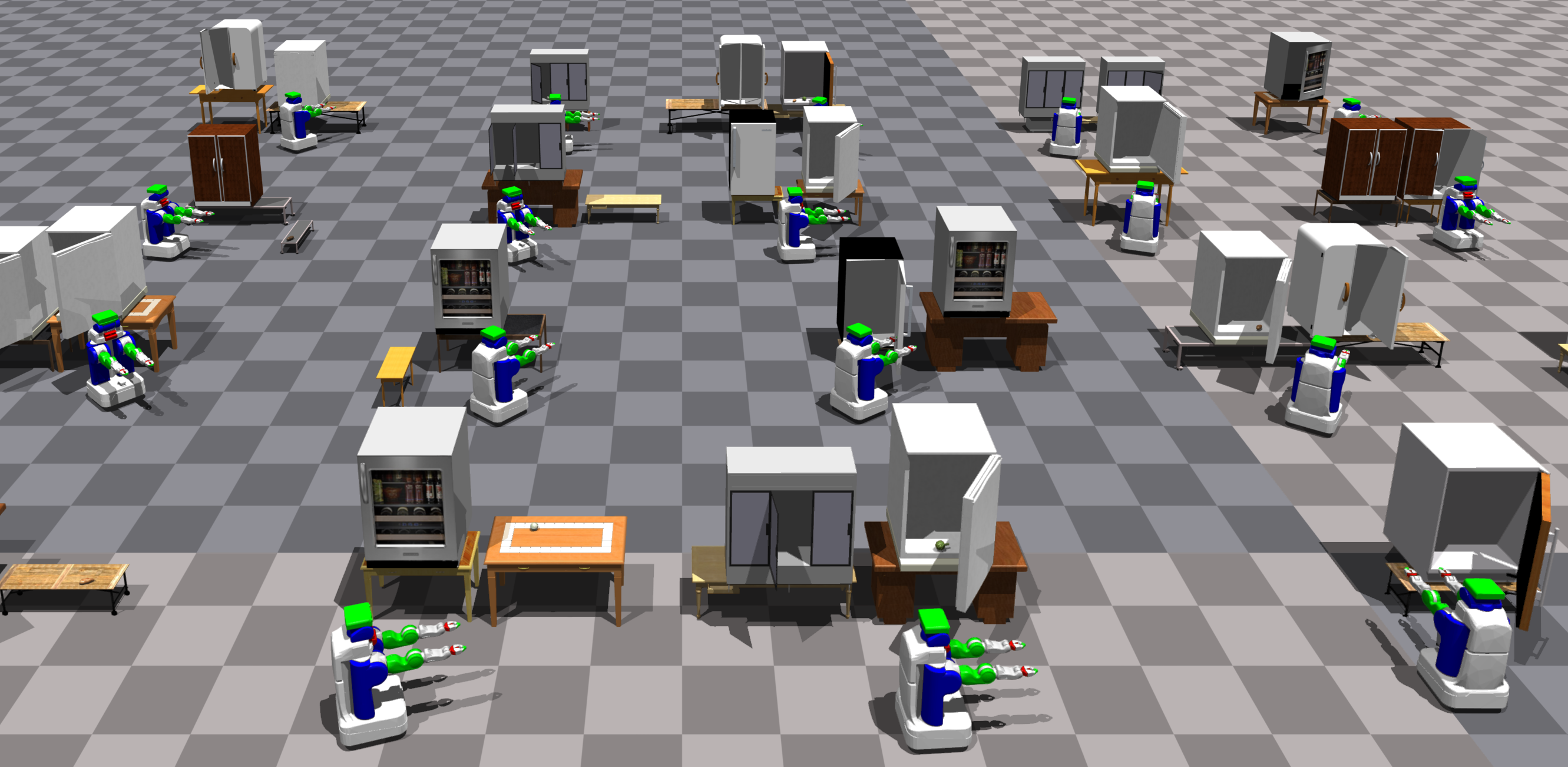}
        \caption{Example scenes in the \textit{Fridges} problem set.}
    \end{subfigure}\hfill 
    \begin{subfigure}[b]{0.49\textwidth}
        \includegraphics[width=\textwidth]{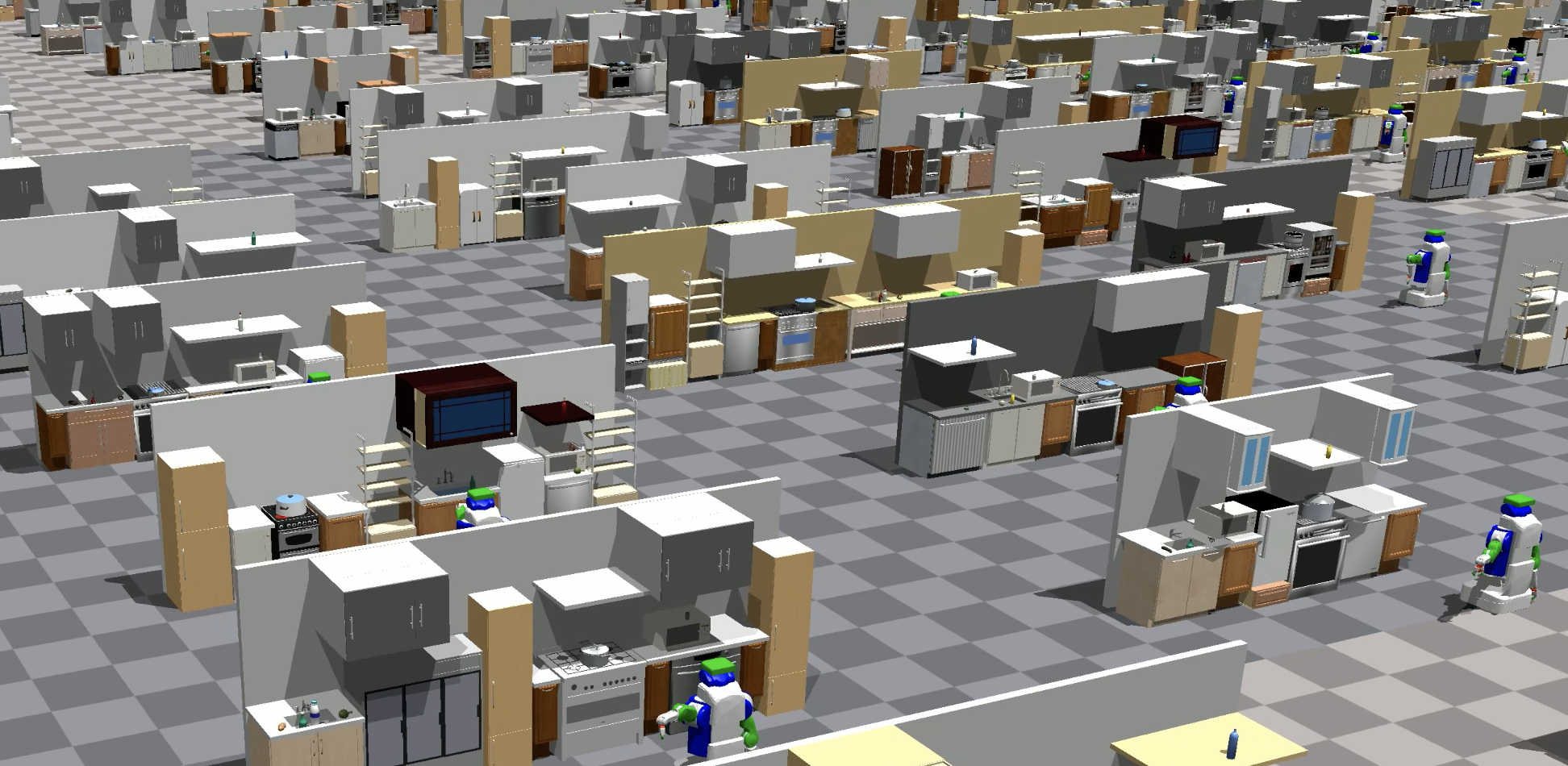}
        \caption{Example kitchen layouts in the Kitchens problem set.}
    \end{subfigure} 
    \caption{Example procedurally generated environment for complex rearrangement problems. To create TAMP problems used for training and testing, we 1) sample a scene, 2) sample a goal, 3) create the initial literals, 4) run a TAMP planner to find a solution, 5) label the task plan associated with the solution as a feasible plan and record the previous task plans that failed due to timeout during refinement as infeasible plans, 6) render segmented RGB images in simulation.}
    \label{fig:scenes}
    \vspace{-15pt}
\end{figure*}


\subsection{The Fridges problem sets} 

There are fridges on top of tables and food items that need to be in the fridge. Each fridge door is either closed 50\% of the time, or open at a position sampled uniformly at random across its joint limits. We also randomly sample the pose of objects, the height of tables, and the initial base configuration of a PR2 robot, as shown in Figure~\ref{fig:scenes}(a). 
We used articulated URDFs from the PartNet-Mobility dataset~\cite{xiang2020sapien} and food meshes sourced online. We generated the scenes using seven fridge assets, nine food assets, and 11 table assets. Overall, the problems contain 1-2 movable objects, 1-2 surfaces, 1-2 storage units, and 1-5 doors. Successful task plans to the problems include 1 pick-and-place and 0-2 pulls.


\subsection{The Kitchens problem sets} 

The \textit{Kitchen} problems have similar randomized properties as the \textit{Fridges} problem set: the poses of movable objects and positions of doors for storage units are randomly sampled. The kitchen environments have different compositions of furniture and appliances but the same number of movable objects, including two food items, two bottles, and two pill bottles in each scene, as shown in Figure~\ref{fig:scenes}(b). The objects are drawn from an asset library consisting of seven assets for each object category, except for the small upper cabinets, of which we used three. The manipulable objects include all goal-related movables and doors, with an additional movable object and one storage unit (with unrelated doors) when the goal region is a storage unit. By adding irrelevant objects to the problem, we test PIGINet's ability to choose plans that don't involve those objects based on the goal and initial state. The other objects in the scene are used as static collision bodies. Overall, the problems contain 2-5 movable objects, 2-4 surfaces, 0-2 storage units, and 0-5 doors. Successful task plans to the problems include 1-3 pick-and-place and 0-2 pulls.

\subsection{Data collection}

After sampling a scene and a goal, we run the {\em batch-sorted} planning algorithm without any feasibility checking to generate up to 100 task plans for the problem. The planner refines each plan until it finds one solution as a positive example. All attempted but failed task plans are labeled as negative examples, as are task plans involving task-irrelevant objects. Labels are noisy, as they are estimates of plan feasibility. Deciding plan feasibility is NP-hard (via a 3-SAT reduction), making obtaining exact labels computationally intractable. Thus, we assign a label to be positive if the planner found a solution within a timeout. We render the segmented images in the PyBullet simulator offline, which includes segmented object links such as doors. If an object is occluded, its associated images will consist entirely of the background color. We save the problem $\langle \mathcal{O}, \mathcal{I}, \mathcal{G}\rangle$ along with corresponding images and one task plan as one data point. 

 We generated a dataset of 600 problems for problem set \texttt{table-to-fridge} and \texttt{fridge-to-fridge}; 500 for \texttt{counter-to-storage} and \texttt{counter-to-pot}, 250 for \texttt{pot-to-storage}, and 150 for \texttt{counter-to-sink} and \texttt{sink-to-storage} each. The names of the problem set describe the initial placement region(s) of goal objects and their destination placement region(s). We both train our models and evaluate the integrated planners on individual problem sets, except for the last one where problems involving two different kinds of goals, \textit{i.e.,} moving to the sink and from the sink, are trained together and tested separately. For training each model, we divide the data with 9-to-1 training/validation split to evaluate training performance and an additional 50 test problems for each task to evaluate the improvement of planning performance.
During training, we augment the images with random crops, rotations, shifts, warps, color jittering, blurring, and grayscale transformations. We used a pre-trained CLIP model~\cite{radford2021learning} for image and text embedding. 
The dataset of problems and solutions will be released upon acceptance of the paper, along with the code for generating kitchen layout.

\begin{figure*}[!tbh]
\begin{center}
\includegraphics[width=1\textwidth]{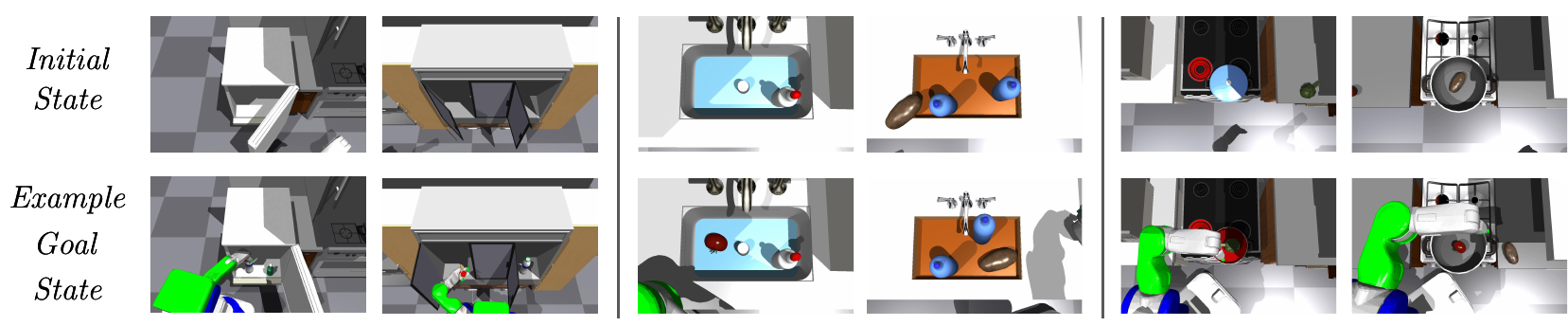}
\end{center}
\caption{Example sources of obstruction in our problem sets. \textbf{Fridges} may have one to three doors that are partially or fully closed. \textbf{Sinks} may be cluttered so object(s) need to be moved away to pick up object from it or place into it. \textbf{Cooking pots} may be occluded by a lid or an object inside, which need to be moved away if there aren't enough space to fit the goal object.}
\label{fig:kitchens}
\end{figure*}

\begin{figure*}[!hbt]
    \centering
    \begin{subfigure}[b]{\textwidth}
        \includegraphics[width=\textwidth]{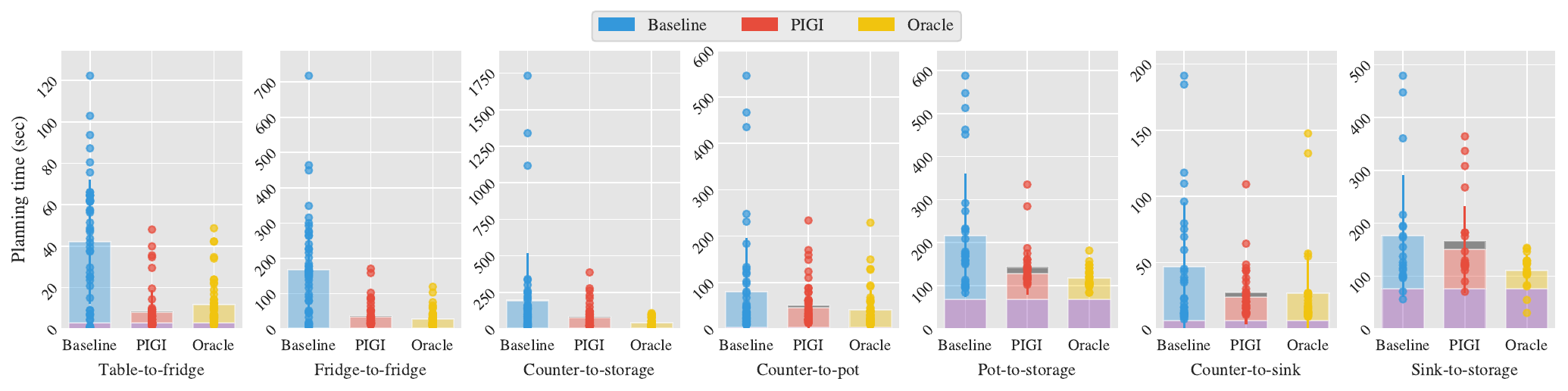}
        \caption{The runtime breakdown of PIGINet-enabled planners compared to a non-learning \texttt{Baseline} and a clairvoyant \texttt{Oracle}. The purple bars indicate average task planning time by FastDownward for generating hundreds of candidate task plans in those problem sets. The gray bars indicate inference time running the candidate task plans through the PIGINet.
        }
    \end{subfigure}\hfill 
    \begin{subfigure}[b]{\textwidth}
        \includegraphics[width=\textwidth]{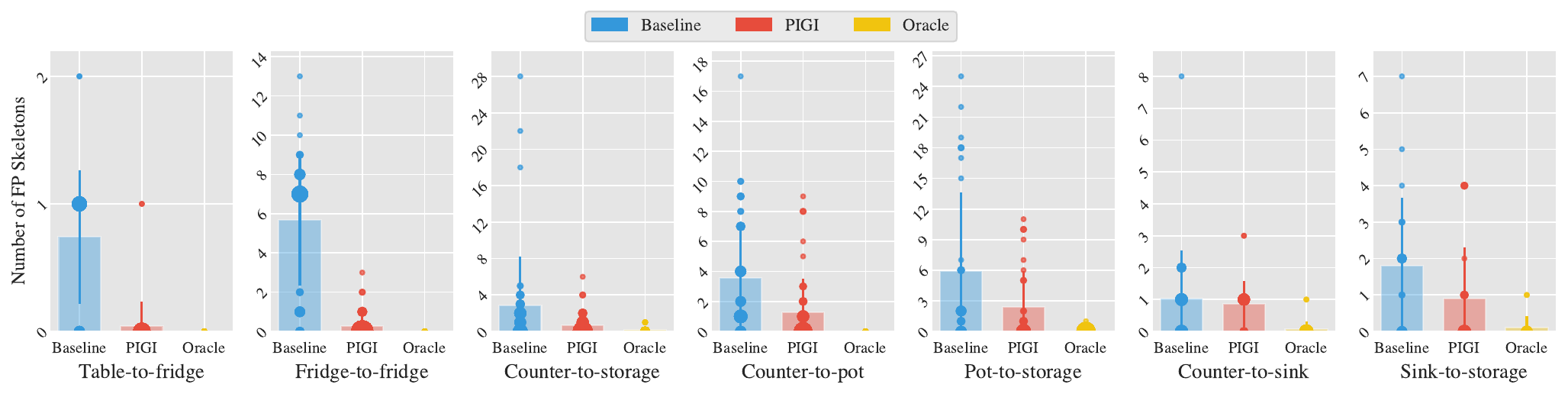}
        \caption{The number of false positive (FP) task plans refined. The size of scattered dots are proportional to density. }
    \end{subfigure} 
    \caption{Evaluation of the PIGINet's ability to reduce planning time, after being trained on individual problem sets and evaluated on 30-50 unseen test problems. Scatter plot points are data points. Bar heights are the mean values with whiskers indicating standard deviation. Each subplot's y-axis is scaled differently. Results show that using PIGINet for feasibility-based plan sorting a) significantly reduces planning time and b) enables the planner to find a solution mostly within three refinement iterations.}
    \label{fig:evaluation}
    \vspace{-15pt}
\end{figure*}



\section{EXPERIMENTS} 

We carried out experiments to answer the following three questions: 
(1) \textbf{Efficiency}: Can PIGINet improve planning speed without sacrificing planner success rate? 
(2) \textbf{Generalization}: Can a trained model make accurate predictions in problems with unseen objects?
(3) \textbf{Ablation}: If we take away parts of the input, can PIGINet still make accurate predictions?

\subsection{Planner performance} 

First, we investigate the effectiveness of PIGINet for speeding up planning. We compared the planning time of three different planner ablations of Algorithm~\ref{alg:batch}. 
1) \texttt{Baseline} is a learning-free planner that always returns $f(\pi, \mathcal{G}, \mathcal{I}) = 1$, attempting to refine every skeleton in the order of ascending plan length.
2) \texttt{PIGI} sorts the plans with the probability generated by our PIGINet. 3) \texttt{Oracle} refines only those task plans that have been logged to be feasible offline, serving as an upper bound on possible performance. 

We test each planner on 30-50 unseen problems for each problem set and record the runtime breakdown as well as the number of infeasible task plans the planner spent on refining before producing a solution. 
We set a 3-60 second timeout for producing diverse task plans using a modified FastDownward planner, depending on the average time it takes to generate a feasible task plan among the batch. We set a 30-60 second timeout on refining each task plan depending on the difficulty of the problem set. We set no total timeout so that all problems are solved by all planners eventually. We set no timeout on batch-producing feasibility scores as they take relatively little time, limited only by the GPU memory available for loading up a trained PIGINet and the number of candidate task plans. As the number of objects and problem horizon increases among problem sets, the number of plans grows exponentially. So does the size of PIGINet, since there are more images from different camera poses and more initial literals to encode. 

Figure~\ref{fig:evaluation}(a) shows that PIGINet is able to cut down planning time across all seven problem sets. It reduced runtime by $80\%$ on the \textit{Fridges} dataset and refinement time by $50\%$ on $Kitchens$ problems. 
Given 50 or 100 task plans, PIGINet usually cuts down the number of infeasible task plans to 1 - 4 and finds a solution after sampling one or two false positive plans, as shown in Figure~\ref{fig:evaluation}(b). From PIGINet's predictions on individual test problems, we observe that it helps with planning the most when there are multiple doors and two of them need to be opened in order to place two objects inside. In those cases, it cuts down dozens of infeasible task plans and ones that the manipulate irrelevant doors. Note that the improvement in the number of false positive plans is larger than that in planning time. 
This disparity is because the shorter plans prioritized by the \texttt{Baseline} planner impose fewer constraints and thus need less time to be proven infeasible during refinement than those longer ones sometimes ranked highly by PIGINet. We think PIGINet's less decisively improved performance on the \textit{Sink-to-storage} is the result of only being able to collect a limited number of training examples (150 problems for each), relative to the large network size required to handle up to 9 movable and articulated objects. 

\subsection{Generalizing to novel objects}

\begin{figure}[tp]
    \centering
    \begin{subfigure}[b]{0.21\textwidth}
        \includegraphics[width=\textwidth]{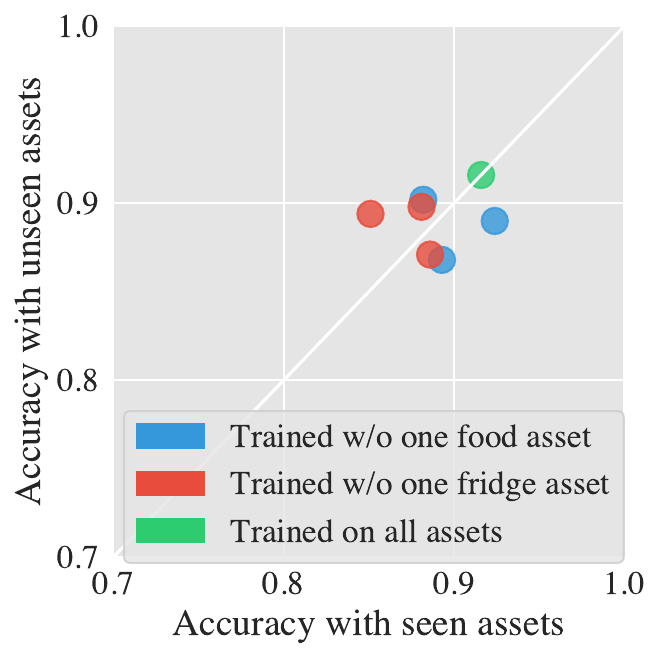}
        \caption{Leave-one-out accuracy.} 
    \end{subfigure}\hfill 
    \begin{subfigure}[b]{0.27\textwidth}
        \includegraphics[width=\textwidth]{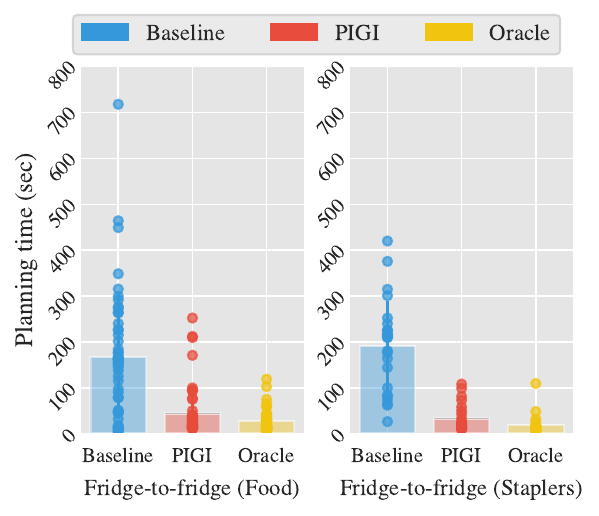}
        \caption{Test runtime for the stapler task.}
    \end{subfigure} 
    \caption{Evaluation of PIGINet's ability to generalization to unseen objects. \textbf{Left:} Each dot indicates one held-out experiment, with x and y values showing the model's prediction accuracy on problems with seen or unseen assets. \textbf{Right:} The planning time reduction by PIGINet is similar when goal objects changed from food to novel category staplers.}
    \label{fig:geometry}
\end{figure}

We show that PIGINet can generalize to problems with unseen objects. For a chosen asset, we held out a test set consisting of problems where the object appeared in the goal condition and train a model with the remaining problems. We compare the model's prediction accuracy on the validation set (with seen assets) and the held-out test set (with unseen assets). We experimented with leaving out three food assets and three fridge assets individually. Figure~\ref{fig:geometry}(a) shows that the models have equally good accuracy on the unseen object assets as the seen ones (scattered dots align around the diagonal line).  We also created a test set where the goal is to move staplers from fridge to fridge. We used five stapler assets from PartNet-Mobility dataset. We see similar improvements in planning time, confirming that PIGINet models can generalize to problems with 
different object categories and shapes. We attribute this impressive generalization ability to 1) the use of a large pre-trained network CLIP and 2) data augmentation techniques used during training, which makes the model less sensitive to colors and texture.

\subsection{Ablation studies}

\begin{figure}[tp]
\begin{center}
\includegraphics[width=0.49\textwidth]{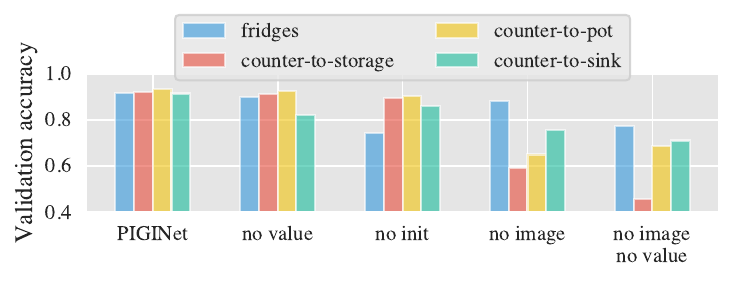}
\end{center}
\caption{Classification accuracy on the validation set across ablations of our method when different input components are removed. Overall, PIGINet taking full input performs the best. }
\label{fig:ablations}
\end{figure}

Finally, we studied how the PIGINet's prediction accuracy is affected when we remove different components of the multi-modal encoding. We compare the models' prediction accuracy on the validation set for four problem sets after the loss converged. The \texttt{fridges} model was trained with both families of problems in the \textit{Fridges} problem sets. Figure~\ref{fig:ablations} shows that PIGINet with all input modalities achieves the best prediction accuracy. Models without continuous values perform almost as well. As for images and initial literals, it seems that discarding initial literals doesn't affect as much when there are a lot of image inputs (due to the 5-6 camera viewpoints used in \textit{Kitchens} problems), compared to the \textit{Fridges} problems where only one image viewpoint is used and objects are scaled differently in the images in order to be sufficiently large. That image cropping process causes PIGINet to lose information about the geometric relationships between objects, which initial literals provide, with literals like \texttt{IsJointTo}(\textit{door1}, \textit{fridge}) and \texttt{SupportedBy}(\textit{Bottle1}, $p_0$, \textit{Sink}). 
These experiments confirmed our hypothesis that images and initial literals contain redundant information as representation of geometric state. This is also good news for using PIGINet for real-robot settings where privileged initial literals that assume full observability can be hard to obtain.






\section{CONCLUSION} 

We developed a novel learning-enabled TAMP algorithm that consists of 1) a multi-modal transformer for predicting feasibility of a batch of candidate task plans given the initial state and goal and 2) a TAMP planner that refines the task plans in the order of predicted feasibility. 
Our method reduces planning time on complex rearrangement problems with articulated and movable obstacles, where uninformed planners suffer from wasted refinement efforts. PIGINet also achieves zero-shot generalization across unseen movable object categories thanks to its visual encoding of objects.



\section*{Acknowledgements}
We would like to thank Weiyu Liu, Wentao Yuan, Valts Blukis, Danfei Xu at NVIDIA Research, and Jiayuan Mao from Learning and Intelligent Systems Group at MIT for helpful discussions on the project. We gratefully acknowledge support from AI Singapore AISG2-RP-2020-016; NSF grant 2214177; from AFOSR grant FA9550-22-1-0249 and from ARO grant W911NF-23-1-0034.



\newpage
\bibliographystyle{unsrtnat}
\bibliography{root.bib}

\newpage
\appendix
\input{appendix}

\end{document}

%% file: appendix.tex
\section{Appendix} 

\subsection{Planning problem sets} \label{app:tasks}

\noindent The \textit{Kitchens} problem sets are as follows:

\begin{enumerate}[label=(\roman*)]  
    \item \texttt{table-to-fridge}: the food is initially on the table. Successful plans involve 1 pick-and-place and 0-2 pulls.
    \item \texttt{fridge-to-fridge}: the food is in another fridge. Successful plans involve 1 pick-and-place and 0-4 pulls.
\end{enumerate}

\vspace{8pt}
\noindent The \textit{Kitchens} problem sets are as follows.

\begin{enumerate}[label=(\roman*)]  \addtocounter{enumi}{2}
    \item \texttt{counter-to-storage}: two instances of food or bottles are to be stored in the fridge or an upper cabinet. Successful plans involve 2 pick-and-place and 0-2 pulls.
    \item \texttt{counter-to-pot}: one food item is to be in the pot. The placement path may be blocked by a lid on the pot and/or an object that's placed inside the pot. The object inside doesn't have to be removed if there is still enough room to fit the target object in. Successful plans involve 1-3 pick-and-place and 0-2 pulls.
    \item \texttt{pot-to-storage}: two food items are to be in the fridge and one of them is initially inside the pot, which may be covered by the lid. Successful plans involve 1-3 pick-and-place and 0-2 pulls.
    \item \texttt{sink}: this is a union of two problem sets involving two different types of goals: (a) \texttt{counter-to-sink}: one food item is to be in the sink, which is occupied by one or two obstacles inside; successful plans involve 1-3 pick-and-place. (b) \texttt{sink-to-storage}: two food items are to be in a storage unit, while at least one of the food item is originally inside the sink with one or two other obstacles; successful plans involve 2-3 pick-and-place and 0-2 pulls. For both sets, the manipulable objects include the target objects, obstacles in the sink, and one extra movable object on the counters. There are 150 problems from each set for training. For evaluating planning time, 30 problems from \texttt{counter-to-sink} and 20 problems from \texttt{sink-to-storage} are used.
\end{enumerate}

\subsection{Model hyper-parameters} 

We used images from five camera poses for problem set (iii, iv) and six camera poses for (v-vi) To efficiently train the transformers, we used a max sequence length of 56 for problem sets (v, vi) and 32 for the others. 





